\documentclass[lettersize,journal]{IEEEtran}
\usepackage{amsmath,amsfonts}
\usepackage{algorithmic}
\usepackage{array}
\usepackage[caption=false,font=normalsize,labelfont=sf,textfont=sf]{subfig}
\usepackage{textcomp}
\usepackage{stfloats}
\usepackage{url}
\usepackage{verbatim}
\usepackage{graphicx}
\usepackage{cite}

\usepackage{multirow}
\usepackage{amssymb}
\usepackage{pifont}
\newcommand{\cmark}{\ding{51}}%
\usepackage{graphicx}
\newcommand\Tstrut{\rule{0pt}{2.6ex}}         
\newcommand\Bstrut{\rule[-0.9ex]{0pt}{0pt}}   
\usepackage{mathtools,amsthm}
\DeclareMathOperator*{\argmax}{arg\,max}
\DeclareMathOperator*{\argmin}{arg\,min}
\usepackage{bm}

\begin{document}

\title{On the Domain Adaptation and Generalization of Pretrained Language Models: A Survey}

\author{Xu Guo, and Han Yu,~\IEEEmembership{Senior Member,~IEEE}
\thanks{Xu Guo and Han Yu are with the School of Computer Science and Engineering, Nanyang Technological University (NTU), Singapore.}
\thanks{Emails: xu008@e.ntu.edu.sg; han.yu@ntu.edu.sg}}



\maketitle

\begin{abstract}
  Recent advances in NLP are brought by a range of large-scale pretrained language models (PLMs). These PLMs have brought significant performance gains for a range of NLP tasks, circumventing the need to customize complex designs for specific tasks. However, most current work focus on finetuning PLMs on a domain-specific datasets, ignoring the fact that the domain gap can lead to overfitting and even performance drop. Therefore, it is practically important to find an appropriate method to effectively adapt PLMs to a target domain of interest. Recently, a range of methods have been proposed to achieve this purpose. Early surveys on domain adaptation are not suitable for PLMs due to the sophisticated behavior exhibited by PLMs from traditional models trained from scratch and that domain adaptation of PLMs need to be redesigned to take effect. This paper aims to provide a survey on these newly proposed methods and shed light in how to apply traditional machine learning methods to newly evolved and future technologies. By examining the issues of deploying PLMs for downstream tasks, we propose a taxonomy of domain adaptation approaches from a machine learning system view, covering methods for input augmentation, model optimization and personalization. We discuss and compare those methods and suggest promising future research directions.
\end{abstract}


\section{Introduction}
\IEEEPARstart{N}{atural}
Language Processing (NLP) is undergoing a paradigm shift with the open-source of large-scale pretrained language models (PLMs), such as GPT \cite{radford2018gpt,radford2019language}, BERT \cite{devlin-etal-2019-bert}, ALBERT \cite{Lan2020ALBERT}, RoBERTa \cite{liu2020roberta}, BART \cite{lewis-etal-2020-bart}, and T5 \cite{Reffel2020T5}. Finetuing PLMs have demonstrated to be a strong baseline on a range of downstream tasks, mostly outperforming previous state-of-the-art task-specific models, making it the de-facto standard in NLP. Having Transformer \cite{NIPS2017_3f5ee243} as their basic neural architecture, these PLMs are trained on broad data using different self-supervised learning tasks at scale. They play a central role as the \textit{foundation model} of AI \cite{bommasani2021-foundation} for their knowledgeable yet incomplete character. The downside of PLMs lies in the lack of portability to real-world domains. PLMs are pretrained by the universal language data, leaving the domain challenges unresolved. A bunch of evidence have surfaced showing that their performance can degrade when they are applied to a narrower domain where data varies substantially from the pretraining corpus \cite{thompson-etal-2019-overcoming,araci2020finbert,chalkidis-etal-2020-legal,miller-etal-2021-domain}. The mismatch between the pretraining and adaptation data distributions makes PLMs struggle to be widely adopted in practice. 

A direct approach to improve generalization performance in machine learning is to enlarge the labeled training data. However, acquiring labels for every task is expensive and time-consuming. To this end, enabling PLMs with domain adaptation (DA) \cite{pan2009survey,weiss2016survey,ramponi-plank-2020-neural,saunders2022domain} which reuses labeled data from related source domains to boost performance on the target domain is necessary. Due to the semantic gap between the embedding spaces of different domains \cite{drift-2021}, directly fitting a single PLM on non-identical domains is suboptimal \cite{bilen2017universal} and may even incur negative transfer due to the domain shifts \cite{lekhtman-etal-2021-dilbert}.  Moreover, performance gains on the target domain can come at the expense of general-domain performance, which is recognized as the catastrophic forgetting problem \cite{mccloskey1989catastrophic}. The effectiveness of domain adaptation largely depends on how well the target domains are represented in the pretraining corpora \cite{radford2019language}. To conduct successful knowledge transfer, a range of domain adaptation methods have been proposed based on different assumptions of the given data and the training setting. However, traditional domain adaptation methods that work with shallow neural networks or non-pretrained language models can be unfavorable to PLMs \cite{ryu2022knowledge}. For example, \cite{wright-augenstein-2020-transformer,karouzos-etal-2021-udalm} find that domain adversarial training on top of BERT is unstable and has little effect on cross-domain performance, suggesting that proper design of domain adaptation methods is necessary for PLMs.

In addition, the long-standing challenge of data scarcity hinders the deployment of many domain-specific systems. Developing domain adaptation and generalization methods for PLMs is promising particularly for data-hungry NLP tasks such as abstractive summarization \cite{yu-etal-2021-adaptsum}. These methods can be incorporated into the pipeline of downstream tasks as easy as playing with "Lego". However, there isn't one-fit-all PLM or on-the-fly domain adaptation methods off the shelf. Through comprehensive review of related literature, we argue that an appropriate choice of domain adaptation methods or a mixture of them can help to avoid performance degradation of PLMs in new domains. For this purpose, this survey aims to provide future researchers and engineers with a road map for accelerating the deployment of PLMs to real-world problems. In the end, we highlight some interesting yet rather meaningful research directions to empower and apply PLMs. 

\noindent\textbf{A Categorization for domain adaptation of PLMs.} We propose a taxonomy (Table \ref{tab:taxonomy}) from a machine learning system view, covering data augmentation, model optimization and personalization techniques. The mainstream approaches focus on data augmentation and model optimization or a combination of them. Personalization has been less explored and most of the methods we surveyed are either recently emerging methods or a resurgence of traditional machine learning approaches. We deem this category is promising for practical adoption of PLMs at scale.

\noindent\textbf{Other surveys.} Comprehensive surveys for domain adaptation or pretrained language models exist, each revisits related works from a different perspective: transfer learning surveys \cite{pan2009survey,weiss2016survey} provide a holistic view including but not limited to DA; DA for visual applications \cite{patel2015visual,csurka2017comprehensive,wang2018deep}; multiple-source domain adaptation (MDA) \cite{mansour2008domain,sun2015survey}; neural UDA for NLP applications based on shallow and non-pretrained language models \cite{ramponi-plank-2020-neural}; DA and MDA for machine translation \cite{saunders2022domain}; taxonomy of PLMs \cite{qiu2020pre} and comprehensive guide to use PLMs for NLP tasks \cite{min2021recent} and particularly for text generation tasks \cite{li2021pretrained}; parameter-efficient adaptation methods for PLMs \cite{ding2022delta}.

\noindent\textbf{Contributions.} Adapting PLMs to specific domains is practically important as PLMs only hold general knowledge. We aim to provide future researchers and engineers with a handbook picturing a systematic view of possible techniques to adapt PLMs to real-world applications more easily. Previous surveys on PLMs focus on the pretraining techniques, surveys on domain adaptation focus on shallow and plain neural networks, and surveys on transfer learning focus on more broad assumptions or settings. In this paper, we systematically examine all the possible solutions for exploiting PLMs to new domains. In addition, we propose a new category named personalization which focus on the setting where the same PLMs are expected to adapt to multiple target domains. We explicitly visualize the assumptions, techniques and PLMs adopted in the literature in Table \ref{tab:visualization} to provide a big picture of the current state of research. Finally, we outline challenges and future research directions.

\section{Preliminaries}
\subsection{Notations}
Throughout the paper, we use $w=\langle w_1,...,w_n\rangle$ to denote a word sequence and $\bm x=\langle x_1,...,x_n\rangle$ to denote their corresponding token embeddings. $y$ denotes the task-specific label for the sequence. $\bm X$ and $Y$ denote the input space and the label space, and the whole dataset is denoted by $\mathcal{D}=(\bm X,Y)=\{(\bm x^{(i)}, y^{(i)}\}_{i=1}^{m}$. Source and target domains are differentiated by the subscriptions $s$ and $t$. E.g., $\mathcal{D}_{s}=(\bm X_s, Y_s)$ denotes a labeled source-domain dataset. We use $p$ to denote a probability distribution and $P$ to denote the probability of an observed event. We use $f_{\theta}$ to denote a PLM parameterized by $\theta$ and $\mathcal{L}$ to denote a loss function. 

\subsection{Theories}
Earlier theoretical \cite{NIPS2006_b1b0432c} and experimental \cite{blitzer-etal-2007-biographies,saenko2010adapting} analysis for domain adaptation demonstrate that the test error of supervised machine learning methods generally increases in proportion to the distribution differences between the training and test sets. 
\cite{ben2010theory}. A theoretical analysis for domain adaptation of language models indicates that pretraining on a larger out-of-domain dataset before fine-tuning on a small in-domain dataset can achieve better generalization than only in-domain training \cite{grangier-iter-2022-trade}. They also show that larger size of pretraining sets does not necessarily bring performance gains to target domains. When their underlying distributions are similar, pretraining can benefit the target-domain tasks.

\subsection{Settings and Assumptions.} 

The rapid development of transfer learning has given birth to a number of transfer learning settings with each holding a different assumption on the given data \cite{pan2009survey}. We review two adaptation settings for PLMs, namely adaptation from pretraining to downstream tasks, which is usually achieved by continual learning (Section \ref{optimization:subsec_1}), and adaptation from related source domains to target domains of our interest. Regarding the feature space and the amount of labels available, recent literature commonly adopt the following assumptions:
\begin{itemize}
    \item[ A1. ] Both domains share the whole feature space. Only the source domain has labels;
    \item[\ A2. ] The source and target domains share a part of the feature space. Only the source domain has labels;
    \item[A3. ] Both domains share the whole feature space. Both domains have labels;
    \item[A4. ] The source and target domains share a part of the feature space. Both domains have labels.
\end{itemize}

We follow \cite{ramponi-plank-2020-neural} and make a difference between supervised domain adaptation (SDA) and unsupervised domain adaptation (UDA) depending on whether the target domain has labels. Note that in existing literature, UDA often assumes a large amount of unlabeled data which is dense while SDA often assumes a small amount of labeled data which is sparse. Therefore, SDA does not always present an easier setting than UDA. A few research papers adopts a semi-supervised domain adaptation setting where a small amount of target-domain data are labeled, which is classified into SDA in our paper.

\begin{table*}[ht]
\begin{center}
  \caption{A taxonomy for the literature on domain adaptation and generalization of PLMs.}
  \label{tab:taxonomy}
  \begin{tabular}{| p{0.18\linewidth} | p{0.42\linewidth} | p{0.28\linewidth} |}
    \hline \Tstrut
    \textbf{Category} & \textbf{Definition} & \textbf{Subcategory} \Bstrut\\
    \hline\Tstrut 
    \multirow{3}{*}{Data Augmentation \ref{sec:data_augmentation}} & \multirow{3}{7.5cm}{Methods that expand or shrink the source-domain training set, or prompt input data with prior information.} & S1: Importance Weighting \ref{data_augmentation:subsec_1} \\  
    & & S2: Psudo Labeling \ref{data_augmentation:subsec_2} \\
    
    & & S3: Prompting \ref{data_augmentation:subsec_3} \Bstrut \\     
     \hline\Tstrut
     \multirow{3}{*}{Model Optimization \ref{sec:optimization}} &
    \multirow{3}{7.5cm}{Methods that optimize the model parameters using different learning objectives to induce better data representations.} & S4: Continual Learning \ref{optimization:subsec_1} \\
    & & S5: Adversarial Learning \ref{optimization:subsec_2} \\  
    
    & & S6: Metric Learning \ref{optimization:subsec_3} \Bstrut \\
    
     \hline\Tstrut
     \multirow{3}{*}{Model Personalization \ref{sec:personalization}} & \multirow{3}{7.5cm}{Methods that adapt the same PLM to multiple different domains together where each domain distribution is relatively stable.} & S7: Posterior Adaptaion \ref{personalization:subsec_1} \\
   & & S8: Specification \ref{personalization:subsec_2} \\
   &  & S9: Reparameterization \ref{personalization:subsec_3} \Bstrut \\
  \hline
\end{tabular}
\end{center}
\end{table*}

\begin{table*}[ht]
\begin{center}
  \caption{A visualization of the assumptions, approaches and PLMs adopted in related work.}
  \label{tab:visualization}
  \begin{tabular}{| p{0.01\linewidth} | p{0.01\linewidth} | p{0.01\linewidth} | p{0.01\linewidth} | p{0.01\linewidth}| p{0.01\linewidth}| p{0.01\linewidth}| p{0.01\linewidth}| p{0.01\linewidth} | p{0.01\linewidth} | p{0.01\linewidth} | p{0.01\linewidth} | p{0.01\linewidth} |  p{0.2\linewidth} |p{0.2\linewidth} |}
  \hline
  
  \multicolumn{4}{|c|}{\textbf{Assumptions}} & \multicolumn{9}{c|}{\textbf{Approaches}} & \multicolumn{1}{c|}{\multirow{2}{*}{\textbf{PLMs}}} &  \multicolumn{1}{c|}{\multirow{2}{*}{\textbf{Related Work}}} \Tstrut\\
     
    \cline{1-13}\Tstrut
     A1 & A2 & A3 & A4 & S1 & S2 & S3 & S4 & S5 & S6 & S7 & S8 & S9 & &  \\
   \hline
    
   \cmark & & & & \cmark &&&&& \cmark &&&&  BERT & \cite{ma-etal-2019-domain} \\\hline
   \cmark & & & & & \cmark &&&\cmark&&&&& ELMo,BERT &  \cite{el-mekki-etal-2021-domain,wang-etal-2019-adversarial} \\\hline 
   \cmark & & & &  &&& &\cmark& \cmark& &&&  BERT,RoBERTa &  \cite{ryu2022knowledge,zou-etal-2021-unsupervised} \\\hline 
   \cmark & & & & & \cmark &&&&\cmark&&&& BERT &  \cite{chen2020adversarial} \\\hline 
   \cmark & & & & & &\cmark &&\cmark&&&&& T5 &  \cite{ye-etal-2020-feature} \\\hline 
   \cmark & & & & & &\cmark &\cmark&&&&&& T5,CPM &  \cite{gu-etal-2022-ppt} \\\hline 
    & & & \cmark & & &&&&\cmark &&&& BERT &  \cite{guo-etal-2021-latent} \\\hline 
     & & & \cmark & & & \cmark &&& &&&& T5 &  \cite{vu-etal-2022-spot} \\\hline 
   \cmark & & & &     & & &   \cmark & & &    & & &     BERT & \cite{karouzos-etal-2021-udalm} \\\hline
   \cmark   & & & &     & \cmark & &    & & &    & & &    BERT  &  \cite{NEURIPS2021_c1fea270}  \\\hline
    \cmark  & & & &     & & \cmark &    \cmark & & &    & & &     T5 &  \cite{gu-etal-2022-ppt}  \\\hline
     \cmark & & & &     & & \cmark &    & \cmark& &    & & &     T5 &  \cite{guo2022improving}  \\\hline
     & & \cmark & &     & & &   \cmark & & &    & & &     BERT,BART,RoBERTa &  \cite{lekhtman-etal-2021-dilbert,rietzler2019adapt,xu-etal-2021-gradual,thompson-etal-2019-overcoming,jin-etal-2022-lifelong}  \\\hline
     \cmark & & & &     & & &   \cmark & \cmark & &    & & &     BERT &   \cite{du-etal-2020-adversarial} \\\hline
     \cmark & & & &     & & &   \cmark & & &    & & &     BERT,BART &  \cite{han-eisenstein-2019-unsupervised,nishida-etal-2020-unsupervised,yu-etal-2021-adaptsum}  \\\hline
    \cmark & & & &     & & &    & \cmark & &    & & &     BERT,RoBERTa &  \cite{lee-etal-2019-domain,wang-etal-2019-adversarial,zou-etal-2021-unsupervised,xie2020unsupervised} \\\hline
     & & \cmark & &     & & &    & \cmark & &    & & &    BERT  &  \cite{park2021consistency}  \\\hline
      & & \cmark & &     & & &    & & \cmark &    & & &     XLM &   \cite{ijcai2020p0508} \\\hline
     \cmark  & & & &     & & &    & & &   \cmark & & &    BERT  &  \cite{guo-et-al-2022-fedhumor}  \\\hline
       & & \cmark & &     & & &    & & &    & \cmark & &     RoBERTa &  \cite{app11051974,chen-etal-2022-modular}  \\\hline
        & & & \cmark &     & & &    & & &    & & \cmark &    GPT-2  &  \cite{chronopoulou-etal-2022-efficient}  \\\hline
        
  \hline
\end{tabular}
\end{center}
\end{table*}

\section{Data augmentation}\label{sec:data_augmentation}

The section of data augmentation includes all the techniques that augment the input data by selecting important samples, adding more useful data samples or adding extra information to adapt the model towards the target domains.

\subsection{Importance Sampling} \label{data_augmentation:subsec_1}

Importance sampling methods \cite{mcbook} identify and select relevant data and try to reduce the negative impact of irrelevant data from source domain during domain adaptation. Earlier research focus on designing a metric or criterion to measure the relevance of a source-domain instance to a target domain with language models as knowledge priors, e.g., the difference between the cross entropy of the sentences from two domains:
\begin{equation}
    \Delta\mathcal{H}(\mathcal{D}_s,\mathcal{D}_t) = \sum_{y\in\mathcal{D}_s}p(y)\mathrm{log}q(y|\theta) - \sum_{y\in\mathcal{D}_t}p(y)\mathrm{log}q(y|\theta),
\end{equation}
where $p$ is the empirical distribution over the domain corpus while $q$ is the distribution predicted by the language model. The top $k$ sentence pairs will be selected to improve model performance by minimizing the importance weighted cross entropy over the source domain dataset $\mathcal{D}_s$:
\begin{equation}
    \mathcal{L}(\theta,\hat{w})=-\frac{1}{|\mathcal{D}_s|}\sum_{y\in\mathcal{D}_s}\hat{w}(y;\mathcal{D}_{s},\mathcal{D}_{t})\mathrm{log}p(y|\theta),
\end{equation}
where $\hat{w}$ estimates the importance weights using metrics such as $\Delta\mathcal{H}$. The quality of sampled training set from the source domain depends on the relative size of the source and target datasets and the quality of the estimators\cite{grangier-iter-2022-trade}. They have been applied to enhance machine translation performance \cite{axelrod-etal-2011-domain,wang-etal-2018-denoising}. 

Dynamic data selection methods \cite{van-der-wees-etal-2017-dynamic} relax the hard selection procedure by assigning the normalized scores to source-domain samples and retain all the source-domain vocabulary while lowering the importance of irrelevant data during training. Influence function \cite{pmlr-v70-koh17a} traces a model's prediction through backpropagated gradients over its training data to identify those training points that are important for making the prediction. It has been applied to the pretrained ResNet for image processing tasks \cite{pruthi2020estimating}, vanilla Transformers for neural machine tranlation \cite{wang2021gradient,mohiuddin2022data,iter2021complementarity} and so on. Other approaches may include training a domain classifier to select source-domain data based on the domain probability \cite{ma-etal-2019-domain}, which involves a multi-source setting. The lower the probability, the more similar the sample is to the target domain. They successfully enhanced domain adaptation of BERT on classification tasks. Importance weighting techniques have also been studied for partial domain adaptation (PDA) where the target-domain classes are only a subset of source-domain classes \cite{zhang2018importance,cao2019learning}. \cite{zhang2018importance} propose a two domain classifier strategy to identify the importance score of source samples. \cite{cao2019learning} propose a progressive weighting scheme to quantify the transferability of source examples to achieve PDA. However, PDA has not been studied for PLMs adaptation yet.

\subsection{Pseudo labeling}\label{data_augmentation:subsec_2}
Pseudo labeling is a straightforward name for methods that use a source-domain classifier to generate pseudo labels for unlabeled data from the target domain. Compared with importance sampling, pseudo labeling focuses on utilizing target-domain unlabeled data. It is also known as self-training \cite{mcclosky-etal-2006-effective} which utilizes the most confident labeled data from target domain to augment the source-domain labeled dataset to continuously train the source-domain model. The resulted model gains an improved discriminative ability on target-domain features. Pseudo labeling can be used in both SDA and UDA settings.

However, pseudo-labels are generally noisy. Since the capacity of PLMs is large enough, simply finetuning PLMs can easily overfit the corrupted labels and therefore hurt the generalization performance. \cite{chen2020adversarial} propose to combine domain-adversarial learning with pseudo labeling where a trainable confusion matrix is optimized against a domain discriminator to reduce the gap between the pseudo-labels and the ground truth.
\cite{el-mekki-etal-2021-domain} applies this approach \cite{chen2020adversarial} to enhance BERT in Arabic cross-domain sentiment analysis. \cite{ye-etal-2020-feature} enhances the quality of pseudo labels by combining self training with knowledge distillation, which distills feature discriminative ability from PLMs to a smaller feature extractor. \cite{NEURIPS2021_c1fea270} propose to reduce the domain shift through cycle self-training where a target classifier is trained on the pseudo labeled target-domain dataset and is required to perform well on the labeled source-domain dataset. Self-training can be extended to gradual domain adaptation in which intermediate domains are treated as target domains step by step \cite{pmlr-v119-kumar20c}. 

Apart from self-training, pseudo labeling can be used to induce domain invariance for domain adaptation. \cite{wang-etal-2019-adversarial} uses a LSTM-based model to generate pseudo questions for target-domain passages and train a domain classifier to discriminate a given passage-question pair as coming from which domain. The answer generator is trained on the induced domain-invariant representations to adapt to the target domain task. Pseudo labeling methods can be applied to solve other machine learning problems such as learning from label proportions \cite{label-dist-2016-ijcai}.

\subsection{Prompting} \label{data_augmentation:subsec_3}
Recently, prompt methods arise as a new paradigm for adapting PLMs to downstream tasks. Prompting refers to methods that prompt the PLMs with additional information about the data such as task descriptions that are used to augment the input data. The resulted input data is usually in a cloze-question format where a template with prompt $w_{p}$ encloses the original instance $w_{in}$ and a mask token is left for PLMs to predict the label words $w_l$:
\begin{equation}
    P(w_{l}|w_{in}, w_{p}) = \argmax_{w_{p}} f_{\theta}(w_{in},  w_{p}).
\end{equation}
With a properly designed prompt for the task at hand, a PLM can correctly generate the label words based on its inner language modeling knowledge priors, which has demonstrated excellent few-shot performance on a range of datasets \cite{brown2020language}. The effect of prompt may be derived from the fact that it leverages the language modeling objective to activate some parameters inside the PLM, making the relationship between the original input data and the label stronger. Recent studies show that the PLMs can behave differently with different kinds of prompts input to GPT \cite{meng2022locating} and other PLMs \cite{creswell2022faithful}.

The additional information that carried by prompts is restricted by the length of manually written prompts. Instead, soft prompt tuning methods \cite{lester-etal-2021-power,li-liang-2021-prefix,liu-etal-2022-p,hambardzumyan-etal-2021-warp} learn prompts through backpropagation on training data. They have demonstrated comparable performance with full-model tuning when the PLMs are large enough. SPOT \cite{vu-etal-2022-spot} proposes to pretrain soft prompts on a set of source-domain datasets and then use the trained soft prompts to boost prompt tuning for target domains. PPT \cite{gu-etal-2022-ppt} introduces unsupervised tasks such as next sentence prediction as the pre-text task for prompt pretraining. After that, the soft prompts are finetuned on the few-shot target-domain data. OPTIMA \cite{guo2022improving} improves over SPOT and PPT by directly performing domain adaptation. Prompts are also shown to boost full-model fine-tuning in LM-BFF \cite{gao-etal-2021-making}, PET \cite{schick-schutze-2021-exploiting,schick-schutze-2021-just}, and PERFECT \cite{karimi-mahabadi-etal-2022-prompt}.

\section{Model Optimization}\label{sec:optimization}
This section describes methods that design loss functions and regularization techniques to adapt models to target domain tasks. We present three subcategories of methods that can achieve this purpose, namely continual pretraining, adversarial learning and metric learning.

\subsection{Continual Learning}\label{optimization:subsec_1}

Continual learning aims to specialize a PLM to a particular domain by continuing the pretraining task on the abundant unlabeled corpora, such as biological documents (BioBERT \cite{lee2020biobert}) scientific papers (SciBERT \cite{beltagy-etal-2019-scibert}), clinical notes (ClinicalBERT\cite{alsentzer-etal-2019-publicly,huang2019clinicalbert}, ClinicalCLNet\cite{huang2019clinical}), financial new (FinBERT \cite{araci2020finbert}), legal documents (LegalBERT \cite{chalkidis-etal-2020-legal}) and tweets (BERTweet \cite{nguyen-etal-2020-bertweet} for English tweets and BERTweetFR \cite{guo2021bertweetfr} for french tweets). During this stage, we continue to train the PLM using the same pretraining task, which is usually a language modeling objective, on the target domain datasets. When data from the target domain are far from enough for the pretraining tasks, then the pretraining objective often plays a role of regularization in the loss functions for the downstream tasks. Based on the purpose of continual pretraining, we categorize methods into vocabulary adaptation, pretrain-finetune and continual adaptation.

\noindent\textbf{Vocabulary Adaptation.} Texts in specialized fields may contain numerous domain-specific terms which are not found in open domain corpora and are therefore not well captured by the vocabulary of PLMs. Domain-specific vocabularies can help domain adaptation of PLMs \cite{gu2021domain}. For example, training the special token embeddings of GPT-2 can enable it in task-oriented dialogue use cases without the need of training new dialogue submodules \cite{budzianowski-vulic-2019-hello}. \cite{zhang-etal-2020-multi-stage} propose to extend the vocabulary of RoBERTa with frequent words from target domains and continue to finetune RoBERTa using a self-constructed reading comprehension task based on coarse annotations. The post-trained RoBERTa was shown to improve low-resource QA tasks from the target domain.  \cite{yao-etal-2021-adapt} propose to expand the vocabulary of BERT with domain-specific corpus and continue to train BERT on the target domain using masked language modeling and knowledge distillation objectives to distill BERT to a small-scale LM which is supposed to be trained as an expert for domain-specific tasks. \cite{sachidananda-etal-2021-efficient} show that the common words from target domains can be represented as the mean of their subword embeddings without further pretraining, which can also effectively adapt BERT to new domains. \cite{poerner-etal-2020-inexpensive} propose to adapt BioBERT to the target domains by aligning its word embeddings to the embeddings trained with Word2Vec \cite{mikolov2013efficient} on the target-domain corpus. To alleviate the semantic shift problem of tokens embeddings during continual pretraining in target domains, \cite{vu-etal-2020-effective} propose a mask learning strategy to adversarially mask out tokens that are hard to reconstruct by the BERT. \cite{lekhtman-etal-2021-dilbert} propose to use aspect category information to selectively mask tokens for masked language modeling and continue to pretrain BERT to induce both domain-invariant and category-invariant representations for cross-domain aspect extraction.

\noindent\textbf{Pretrain-Finetune.} Studies in  \cite{gururangan-etal-2020-dont} show that it is helpful to tailor a pretrained model to the domain of a target task through a second phase of pretraining for both high- and low-resource settings. The cross-domain adapted BERT \cite{rietzler2019adapt} demonstrates that domain-specific language modeling followed by supervised task-specific finetuning can significantly boost aspect-based sentiment classification. \cite{karouzos-etal-2021-udalm} propose to add MLM loss on the target-domain data as regularization during finetuning BERT for source-domain sentiment classification. \cite{du-etal-2020-adversarial} propose to enable BERT with domain awareness by introducing adversarial domain discrimination into the continual pretraning stage where BERT is further pretrained on the sentiment classification datasets using masked language modeling. Unsupervised domain adaptation of BERT to languages of special genres, such as Early Modern English and Tweets, can be achieved by only finetuning the contextualized embeddings using masked language modeling on unlabeled text from the target domain \cite{han-eisenstein-2019-unsupervised}. Continual pretraining on unlabeled data from target domain using language modeling has shown to be effective to adapt AraBERT \cite{antoun-etal-2020-arabert} to tweets data for arabic dialect identification \cite{beltagy-etal-2020-arabic}. \cite{nishida-etal-2020-unsupervised} shows that finetuning BERT with language modeling on the target-domain datasets while performing reading comprehension on the source-domain QA datasets can better adapt BERT to the target-domain QA datasets. 

\noindent\textbf{Continual Adaptation.}  Continual pretraining is closely related to lifelong learning or continual adaptation \cite{parisi2019continual} in which a general model is continuously adapted to new domains. \cite{xu-etal-2021-gradual} shows that gradually finetuning BERT-based dialogue models in a multi-stage process is better than one-stage finetuning. \cite{thompson-etal-2019-overcoming} adopt the lifelong learning setting and train the BART across different domains for text generation. Despite simple and easy to deploy, this learning setting typically incur a catastrophic forgetting problem \cite{mccloskey1989catastrophic,kirkpatrick2017overcoming}. Blindly continue pretraining a given PLM on the target domain can be trapped in the forgetting problem. Studies in \cite{yu-etal-2021-adaptsum} demonstrate that the dissimilarity between the pretraining data and target domain task can degrade the effectiveness of BART in abstractive summarization in which case seeking a relate source domain to perform domain adaptation can be helpful. The performance degradation on the target domain can be reduced by inventing more advanced techniques such as look-ahead learning on the domain discriminator under adversarial neural transfer \cite{guo-etal-2021-latent}, where BERT representations can be better adapted to the target domain. To enable temporal domain adaptation of PLMs to emering data, \cite{jin-etal-2022-lifelong} studied different continual learning algorithms to continue pretraining RoBERTa in new domains. Experiments show that distillation-based continual learning achieves better temporal generalization performance than other possible solutions include tuning domain-specific adapters \cite{adapter-houlsby19a-icml} and memory replay methods \cite{chaudhry2019tiny}.

\subsection{Adversarial Learning}\label{optimization:subsec_2}
Adversarial learning methods generally employ a GAN-like setup \cite{NIPS2014_5ca3e9b1} where a domain discriminator is optimized against the task-specific learning objectives. 

\noindent\textbf{Domain-adversarial Training.} Instead of directly fitting a single PLM on non-identical domains, the leading solution to this problem is to reconfigure the network into domain-agnostic and domain-specific layers \cite{rebuffi2017learning,wang2019towards}. The mainstream domain adaptation approaches in the literature are developed based on domain-adversarial neural networks (DANN) \cite{ganin2016domain} or adversarial discriminative domain adaptation (ADDA) framework \cite{tzeng2017adversarial}. The goal is to induce domain-invariant representations via the domain-agnostic layers and map the source and target data into a common feature space by solving a min-max game between
\begin{equation}
    \argmin_{\theta_{G}}\mathcal{L}_{C}(x_s, y_s) - \mathcal{L}_{AD}(x_s, x_t),
\end{equation}
and
\begin{equation}
    \argmin_{\theta_{D}}\mathcal{L}_{AD}(x_s, x_t),
\end{equation}
where $\theta_{D}$ denotes the parameters of domain discriminator and $\theta_{G}$ denotes the reset parameters of the model including the task classifier. $\mathcal{L}_{AD}$ computes the cross entropy for domain classification over source and target domains:
\begin{equation}
    \mathcal{L}_{AD}(x_s, x_t) = - \mathrm{log}(P(x_s=1)) - \mathrm{log}(P(x_t=0)).
\end{equation}
\cite{lee-etal-2019-domain} employs a domain discriminator and applies domain-adversarial training to achieve domain generalization of BERT for QA tasks. \cite{wang-etal-2019-adversarial} generates pseudo questions for unlabeled target-domain passages and a domain classifier is applied on top of BERT to discriminate which domain a passage-question pair comes from. \cite{zou-etal-2021-unsupervised} use the domain discriminator to deceive an autoencoder to enforce RoBERTa to produce domain-invariant representations. \cite{ghosal-etal-2020-kingdom} propose to improve DANN with an external knowledge base, ConceptNet, to enhance both domain-specific and general knowledge extraction for cross-domain sentiment analysis. \cite{tang2020unsupervised} propose to exploit structural domain similarity to enhance the discriminability of domain-invariant representations for the target-domain data. 

Studies in \cite{guo-etal-2021-latent,ryu2022knowledge} found that a catastrophic forgetting problem occurs when the ADDA framework is applied to the BERT model. \cite{guo-etal-2021-latent} propose a look-ahead optimization strategy to accommodate the adversarial domain discrimination loss and the task-specific classification loss when optimizing BERT representations. \cite{ryu2022knowledge} propose to use knowledge distillation \cite{hinton2015distilling} to distill knowledge from source encoder to target encoder, thereby regularizing ADDA for unsupervised domain adaptation of BERT.

\noindent\textbf{Adversarial Robustness and Consistency Training.}
Adversarial robustness refers to ensuring models to be robust against adversarially generated perturbations. Consistency training \cite{NIPS2016_30ef30b6,xie2020unsupervised} forces the model to make consistent predictions against small perturbations which are not necessarily to be adversarial noise. Both techniques try to smooth the decision boundary to improve the generalization performance of a model in the face of a small distribution deviation within a tolerance bound:
\begin{equation}
    x_s^\prime = x_s + \epsilon \bigtriangledown_{x_s}\mathcal{L}_{C}(x_s, y_s).
\end{equation}
The perturbed samples $x_s^\prime$ will be added to the training set to train the model to minimize the original classification loss over them:
\begin{equation}
\mathcal{L}_{AT} = \mathcal{L}_{C}(x_s, x_s^\prime, y_s)     
\end{equation}

This kind of regularization technique has been widely adopted in NLP. For example, \cite{park2021consistency} produce discrete virtual adversarial noise to the token embeddings. \cite{yoon-etal-2021-ssmix} apply mixup to perturb the spans of the input texts for text classification for consistency training. \cite{kim-etal-2021-learn} propose a consistency training framework to enhance the conversational dependency of question answering. They have shown to be able to boost the generalization performance of a model. Recent studies show that AT can also help domain adaptation by focusing on smoothing the decision boundary where source and target domain are similar \cite{pmlr-v97-liu19b,guo2022improving,jiang2020bidirectional}:
\begin{equation}
    \mathcal{L}_{all} = \mathcal{L}_{C}(x_s, y_s) + \lambda_{1}\cdot\mathcal{L}_{AT}(x_s, y_s) + \lambda_{2}\cdot\mathcal{L}_{AD}(x_s, x_t).
\end{equation}
Using $\lambda_{AD}$ to generate perturbations can reduce the domain gap thereby enhancing domain adaptation \cite{jiang2020bidirectional}. In \cite{guo2022improving}, authors found that optimizing the domain discrimination loss and task classification loss for T5-based prompt tuning across domains suffer from low capacity of the soft prompts while applying $\lambda_{AD}$ to generate transferable perturbations can avoid the loss competition problem \cite{guo-etal-2021-latent}. Moreover, the problem of tail classes alignment across domains can also be alleviated by training against adversarial perturbations for semantic segmentation \cite{yang2020adversarial}. However, this topic has not been studied in NLP yet. 

\subsection{Metric Learning}\label{optimization:subsec_3}
Metric learning techniques have also been explored for the purpose of domain adaptation. The goal of applying metric learning is to train the neural networks to optimize a designed metric such that the resulted representations can have certain property. Earlier research focus on aligning the output distributions of the source and the target domains by minimizing the discrepancy between them. \cite{tzeng2014deep} was the first to adopt Maximum Mean Discrepancy (MMD) \cite{gretton2012kernel} metric for both SDA and UDA settings. MMD is computed on the CNN representations of source and target images as a measurement of distribution discrepancy:
\begin{equation}
    \mathcal{L}_{\mathrm{M}}=\lVert \frac{1}{|X_s|}\sum_{x_s\in X_s}f_{\theta}(x_s) - \frac{1}{|X_t|}\sum_{x_t\in X_t}f_{\theta}(x_t) \rVert.
\end{equation}
The model tries to learn representations that are invariant to source and target domains by minimizing the squared MMD loss together with the task-specific loss:
\begin{equation}
\mathcal{L}=\mathcal{L}_C(X_{l}, Y) + \lambda \mathcal{L}_{\mathrm{M}}^{2}(X_s, X_t),
\end{equation}
where $X_{l}$ contain all the labeled data from source and target domains. Another commonly used metric is correlation analysis. \cite{sun2016deep,sun2016return} was the first to propose to use correlation analysis to reduce the domain shift in UDA. \cite{rahman2020correlation} combine the correlation analysis and adversarial learning to achieve domain adaptation and generalization. The goal of correlation-based domain alignment is to minimize the difference between the covariance of the source features and the covariance of the target features:
\begin{equation}
\mathcal{L}_{cor}(x_s, x_t) = \lVert \mathrm{cov}(f_{\theta}(x_s)) -\mathrm{cov}(f_{\theta}(x_t)) \rVert^{2}_{F},
\end{equation}
which is used to regularize the overall training objective: 
\begin{equation}
    \mathcal{L}=\mathcal{L}_C(X_{l}, Y) + \lambda \mathcal{L}_{cor}(X_s, X_t). \\
\end{equation}

Mutual information (MI) has also been exploited for domain adaptation \cite{ijcai2020p0508} in which the MI between the representations from two domains are maximized to extract domain-invariant features on top of XLM \cite{conneau2019cross}. This kind of approach stems from Informax optimization, which refers to the principle that when a set of input values is mapped to a set of output values through a function, the Shannon mutual information between them should be maximized. Enforcing neural network representations of data to match a specific statistical prior came with adversarial autoencoders \cite{makhzani2015adversarial}. Deep Informax \cite{hjelm2018learning} extends this idea to Informax Optimization problems to constrain representation learning. The quality of the learned representations can be measured by the mutual information between them and the corresponding input data \cite{mutualinfo-belghazi-2018-icml}. They can be used for independent component analysis.

\section{Personalization}\label{sec:personalization}

Understanding personal habits of language usage in terms of named entities \cite{li-etal-2018-named},
part-of-speech \cite{sennrich-haddow-2016-linguistic}, and syntactic structure \cite{aharoni-goldberg-2017-towards}, is important to personalize a system to different users. These information are contained in domain-specific data. The challenge is that finetuning every copy of the same PLM on a different domain could be prohibitive as the model size and the number of domains grow. The task of personalizing the same PLMs to different domains at scale is at the intersection between domain adaptation and personalized federated learning (PFL) \cite{tan2022towards}. The latter is proposed to solve the problem in which FL-trained models incur performance drop across different data distributions from different clients. In view of the practical significance of providing customized NLP service for applications such as personalized response generation \cite{response-sigir-2017,reposnse-per-2022}, we aggregate those methods that are promising to adapt the same PLMs to different domains into this category. There are three kinds of ways to achieve this. Posterior adaptation methods study how to adapt a fixed pretrained model to a domain where the label distributions shift from the training data. Specification methods specifies a small amount of the inner parameters of PLMs to be tuned using the domain-specific datasets. Reparameterization methods inject new parameters to PLMs without changing any of the pretrained parameters.

\subsection{Posterior Adaptation} \label{personalization:subsec_1}
Empirical rish minimization trains a neural model to estimate the \textit{posterior} probability $\hat{p}(Y|X)$ to describe how likely the observed training data $x$ happen to be the label $y$. The model tries to approximate the true class priors $p(y|x)$ by learning from more representative training data or using different optimization techniques. However, in practical evaluation scenarios, the prior probabilities may differ from that of the training set and may even change from one domain to another, which is often called prior shift or label shift \cite{vsipka2022hitchhiker}. Coping with prior shift is important for personalizing PLMs to multiple different domains since re-training a PLM is quite expensive and can easily overfit an imbalanced and small dataset. Based on Bayes rules, we can derive the following:
\begin{equation}
    p(y|x)=\frac{p(y)}{p(x)}\cdot\frac{\hat{p}(y|x)\hat{p}(x)}{\hat{p}(y)}\propto \hat{p}(y|x)\cdot\frac{p(y)}{\hat{p}(y)},
\end{equation}
where the ratio $\frac{p(y)}{\hat{p}(y)}$ implies the prior shift. In \cite{vsipka2022hitchhiker}, authors propose the test-time adaptation of a fixed pretrained classifier after a prior shift happens by re-weighting its predictions based on confusion matrices. To avoid over-confident predictions due to overfitting to some classes, \cite{alexandari2020maximum} propose to calibrate the confidence of classifier predictions by adding class-specific bias terms:
\begin{equation}
    \hat{p}(y|x)=\frac{\mathrm{exp}(z_i(x)/\beta + b_i)}{\sum_{j}\mathrm{exp}(z_j(x)/\beta + b_j)},
\end{equation}
where $z_i(x)$ represents the output logits of the input $x$ and $\beta$ is a temperature scaling factor. These methods focus on solving the label shift problem of one target domain. Extending this problem to multiple domains calls for another line of machine learning research called Learning from label proportions (LLP) in which the training data is provided in groups and only the label distribution for each group is given \cite{label-proportion-2008-icml,label-proportion-2010-icml}. Given a model parameterized by $\theta$, the task is to predict individual labels $y\in\{-1, 1\}$ for each group. The key is how to utilize the given label proportion to optimize model's predictions. \cite{pmlr-v28-yu13a} propose the $\varpropto$SVM regularization approach which minimizes a penalization term $\mathcal{L}_{pn}$ to reduce the difference between the true label proportion $p_{k}$ and the estimated label proportion $\hat{p}_{k}$ of group $k$ :
\begin{equation}
    \mathcal{L}_{pn}(\hat{p}_{k}(y|x), p_{k}(y); \theta) = \lvert \hat{p}_{k}(y|x) - p_{k}(y) \rvert
\end{equation}
Solving LLP problems enables interesting applications such as modeling voting behaviors across different demographic groups \cite{pmlr-v28-yu13a}. However, the use of LLP methods may raise concerns about privacy leakage resulted from observing label proportions. To mitigate this issue, \cite{guo-et-al-2022-fedhumor} propose to adopt federated learning framework in which training data and label proportions are kept on local devices while only the model parameters are communicated between devices. On each client the true label proportion is used to penalize the estimated one with a temperature scaling factor $\beta$ tuned on validation set:
\begin{equation}
    \mathcal{L}_{pk}(\hat{p}_{k}(y|x), p_{k}(y);  \theta)= \frac{\hat{p}_{k}(y|x)}{p_{k}(y)^{\beta}}
\end{equation}

\subsection{Specification}\label{personalization:subsec_2}
Specification methods specify a part of the parameters of a PLM to be tuned for domain adaptation. A cross-lingual study \cite{app11051974} found that selectively post-train parameters of RoBERTa which is pretrained on high-resource languages can better adapt it to low-resource languages.  \cite{Michel2018ExtremeAF} propose personalized model adaptation solely performed on the output vocabulary bias vector. \cite{wuebker-etal-2018-compact} propose a parameter-efficient domain adaptation setting for training personalized machine translation models. Most of the model parameters are frozen during training while a set of offset tensors are personalized to each user and need to be trained. Structured sparsity is encouraged on the offset tensors via group lasso regularization \cite{group-lasso-2017} to reduce parameters consumption from each user. Similar as the design of residual adapters for visual domains \cite{rebuffi2017learning}, modular domain adaptation for text understanding \cite{chen-etal-2022-modular} also successfully applied domain-specific bias and normalization terms in customizing models to different users. \cite{NEURIPS2021_PFL} proposes partial PFL which loads a subset of the global model's parameters as initialization on each client and shows improved generalization under cross-domain evaluation. BitFit \cite{zaken-2022-bitfit} updates the bias of PLMs while freezing the rest parameters. Their ablation studies also show that finetuning only a subset of all the bias terms in the PLM can achieve similar performance as finetuning the whole bias set, indicating a specialization on the bias set is promising to adapt the PLM to multiple domains together. However, when equipped with large enough PLMs, different parameter-efficient adaptation methods result in similar performance \cite{ding2022delta}, indicating an upper bound may exist for the adaptation performance of PLMs. 

It is possible to localize knowledge in a PLM in order to make targeted parameter updates without forgetting most of the already-learned knowledge \cite{dai-etal-2022-knowledge,de-cao-etal-2021-editing,mitchell2022fast}. However, repeated editing the PLM can still exhibit forgetting problems \cite{hase2021language}. Elucidating the scenarios in which personalization does or does not benefit performance is an important direction for future work.

\subsection{Re-parameterization}\label{personalization:subsec_3}
Re-parameterization methods adapt large-scale PLMs by optimizing a low-dimensional subspace of the model or transformed from the model. Adapter modules \cite{rebuffi2017learning,wang2019towards} come to compress many visual domains together to adapt a single model to multiple domains together without ignoring domain-specific features. The first approach to adapt a frozen PLMs to domain-specific datasets came with Adapter tuning \cite{adapter-houlsby19a-icml} which inserts an adapter module, simply two linear layers with activation and skip-connection, between transformer blocks of PLMs. \cite{cooper-stickland-etal-2021-multilingual} injects domain-specific and language-specific adapters to a vanilla Transformer which is pretrained on multilingual data and adapt it to new domains and new languages together for machine translation. \cite{chronopoulou-etal-2022-efficient} specializes GPT-2 in a number of domains by constructing a hierarchical tree with each node associated with an adapter module. The GPT-2 is finetuned together with those adapters with the task of language modeling. The hierarchical adapters allows the partially sharing of similar domains, which generalizes better than assigning each domain with a domain-specific adapter and enforcing all domains to share the same adapter. AdapterDrop \cite{ruckle-2021-adapterdrop} enhances the generalization performance of Adapter tuning by learning to drop out some adapter layers. Compacter \cite{mahabadi2021compacter} reduces trainable parameters of adapter by decomposing the linear layers into low-rank matrices while maintaining the same performance. Parallel Adapter \cite{he2022-paralleladapter} inserts an adapter to every transformer layer in parallel which allows adaptation to be faster than the sequential Adapter \cite{adapter-houlsby19a-icml}. LoRA \cite{hu2022lora} injects a trainable low-rank matrix aside each dense layer in the transformers to enforce the layer parameters to be decomposed into low-rank matrices.

\subsection{Evaluation Benchmarks}
Earlier cross-domain evaluation benchmarks mainly use the cross-domain aspect extraction \cite{jakob-gurevych-2010-extracting} and multi-domain sentiment analysis on Amazon reviews \cite{blitzer-etal-2007-biographies}. Named entity recognition (NER) datasets include \cite{salinas-alvarado-etal-2015-domain} for finance domain, WNUT2016 \cite{strauss-etal-2016-results} for tweets genre, and BioNLP \cite{kim-etal-2009-overview,kim-etal-2011-overview} for biomedical domain. AdaptSum \cite{yu-etal-2021-adaptsum} simulates the low-resource setting for the abstractive summarization task with a combination of existing datasets across six diverse domains. MultiWOZ \cite{budzianowski-etal-2018-multiwoz} is a collection of human-tohuman conversation transcriptions in multiple domains. \cite{budzianowski-vulic-2019-hello,xu-etal-2021-gradual} are evaluated on MultiWOZ.

\section{Future Research Directions}

\subsection{Low-resource Learning}

Domain adaptation enables data-efficient learning with PLMs by transferring knowledge from high-resource to low-resource domains. \cite{chen-etal-2020-low} adopted meta learning for low-resource domain adaptation of BART in task-oriented semantic parsing. They argue that better representation learning and better training techniques are crucial for adapting PLMs to low-resource tasks. \cite{guo-etal-2021-latent} show that optimizing over BERT representations instead of the entire model parameters can alleviate over-fitting high-resource domains when adapting to low-resource domains. 

A range of parameter-efficient adaptation methods has shown to be an alternative to finetuning for PLMs. With a few parameters tuned, the performance on downstream tasks can be comparable with finetuning \cite{adapter-houlsby19a-icml,lester-etal-2021-power,devlin-etal-2019-bert}. 
The rising research on parameter-efficient methods for PLMs adaptations and some pioneering transfer learning studies under this sector \cite{guo2022improving,gu-etal-2022-ppt} show the promise of low-recourse learning with PLMs. However, despite exciting results, little has been explored about critical ingredients and mechanisms that made these methods work. On one hand, studies in \cite{he-etal-2021-effectiveness} show that Adapter-based tuning can better mitigate the forgetting problems than finetuning on low-resource and cross-lingual tasks. On the other hand, studies in \cite{guo2022improving,gu-etal-2022-ppt} show that prompt tuning under-performs finetuning in few-shot settings. These evidence suggest that more attention need to be paid to understand the behavior of different parameter-efficient adaptation methods in low-resource learning.

\subsection{Interactive and Lifelong Learning}

PLMs have demonstrated to be knowledgeable AI artifacts and they have shown enough capacity to keep lifelong learning \cite{PARISI201954,thompson-etal-2019-overcoming}, which can be treated as an extreme example of domain adaptation. In real-world scenarios, PLMs are expected to interact with human and keep updated to new domains without forgetting the general knowledge. Continual finetuing of PLMs from one domain to another in a sequential manner can be easily trapped in the catastrophic forgetting problem. Elastic Weight Consolidation \cite{kirkpatrick2017overcoming} is an approach to mitigate the problem for lifelong learning. It works by only allowing model parameters that are less important to general-domain performance to be adapted to the target domain. The idea is coincidentally similar to the recently rising research on parameter-efficient adaptation of PLMs, e.g., prompt tuning\cite{lester-etal-2021-power}, bears the similar idea. On top of frozen PLMs, prompt tuning shows competitive generalization performance as full-model finetuning. With most of the prelearned parameters untouched, PLMs requires only a few parameters to be trained. The use of some properly designed domain adaptation methods can further boost parameter-efficient learning with a few examples \cite{guo2022improving}, which is promising to circumvent the episodic memory requirement in lifelong learning \cite{kirkpatrick2017overcoming,chaudhry2019tiny}.

\subsection{Customized NLP Service}

The diversity among different users can hinder the acceleration of scaled product service provided by PLMs. Personalization is critically important to empower interactive systems such as virtual assistants where a single model is expected to be adapted to many users based on the understanding the user's intents. Domain adaptation techniques are promising to achieve this purpose. Domain adaptation for personalization has been well studied in CV \cite{per-facial-2014,ahuja2022low} and signal processing \cite{stress-per-2018,ECG-per-2019}. In NLP, personalization has been mainly studied in customized machine translation \cite{denkowski-etal-2014-learning,peris2017online,turchi2017continuous,li-etal-2018-named}. Applying personalization methods such as posterior adaptation allows incremental adaptation of PLMs in which a user can provide a correct translation of each segment just after receiving machine translation suggestions and the system is able to train on that correction before generating next suggestions \cite{denkowski-etal-2014-learning,peris2017online,turchi2017continuous}.


\bibliography{custom}
\bibliographystyle{IEEEtran}












\newpage

 




\vfill

\end{document}